\documentclass[lettersize,journal]{IEEEtran}
\usepackage[Symbolsmallscale]{upgreek}
\usepackage{xargs}
\usepackage{float}
\usepackage{wrapfig}
\usepackage[normalem]{ulem}
\usepackage{hyperref}
\usepackage{amsmath,amsfonts}
\usepackage{algorithm}
\usepackage{array}
\usepackage[caption=false,font=normalsize,labelfont=sf,textfont=sf]{subfig}
\usepackage{textcomp}
\usepackage{stfloats}
\usepackage{url}
\usepackage{verbatim}
\usepackage{graphicx}
\usepackage{cite}
\usepackage{hyperref}
\usepackage{xcolor}

\usepackage{multirow}
\usepackage{amsthm}
\usepackage{bbm}

\usepackage{graphics} 
\usepackage{epsfig} 
\usepackage{mathptmx} 
\usepackage{times} 
\usepackage{amsmath} 
\usepackage{amssymb}  
\usepackage{textcomp, gensymb}
\usepackage{url}
\usepackage{graphicx}
\usepackage{color}
\usepackage{xcolor}
\usepackage{soul}
\usepackage{mathtools}
\usepackage{algorithm, setspace}
\usepackage{svg}
\usepackage[noend]{algpseudocode}
\usepackage{ccicons}
\usepackage{comment}
\newtheorem*{lemma*}{Lemma}

\newtheorem*{remark}{Remark}
\usepackage{colortbl}

\newcommand{\state}{\mathbf{x}}
\newcommand{\dyn}{f}

\newcommand{\ctrl}{u}

\newcommand{\obs}{\mathcal{O}}

\newcommand{\env}{E}
\newcommand{\envp}{d}
\newcommand{\policy}{\uppi}
\newcommand{\sensor}{S}
\newcommand{\image}{I}

\newcommand{\NRT}{NRT}
\newcommand{\VBC}{VBC}

\newcommand{\traj}[2]{\zeta^{#1}_{#2}}

\newcommand{\brt}{\mathcal{V}}
\newcommand{\ibrt}{\mathcal{I}_{unsafe}}
\newcommand{\inprod}[2]{{\left\langle #1, #2 \right\rangle}}
\newcommand{\classy}{\sigma}
\newcommand{\confclassy}{\hat \sigma}
\newcommand{\dr}{V_\theta}

\begin{document}

\title{Enhancing Safety and Robustness of Vision-Based Controllers via Reachability Analysis}

\author{Kaustav Chakraborty$^{1}$, Aryaman Gupta$^{1}$, Somil Bansal$^{1,2}$

\thanks{$^{1}$Authors are with the ECE Department at the University of Southern California. {\tt\footnotesize \{aryamang, kaustavc, somilban\}@usc.edu.}}
\thanks{$^{2}$Author is with the AA Department at Stanford University.}
\thanks{This research is supported in part by the DARPA Assured Neuro Symbolic Learning and Reasoning (ANSR) program and by the NSF CAREER program (2240163).}}

\markboth{Journal of \LaTeX\ Class Files,~Vol.~14, No.~8, August~2021}%
{Shell \MakeLowercase{\textit{et al.}}: A Sample Article Using IEEEtran.cls for IEEE Journals}

\maketitle

\begin{abstract}
Autonomous systems, such as self-driving cars and drones, have made significant strides in recent years by leveraging visual inputs and machine learning for decision-making and control. Despite their impressive performance, these vision-based controllers can make erroneous predictions when faced with novel or out-of-distribution inputs. Such errors can cascade into catastrophic system failures and compromise system safety.
In this work, we compute Neural Reachable Tubes, which act as parameterized approximations of Backward Reachable Tubes to stress-test the vision-based controllers and mine their failure modes. The identified failures are then used to enhance the system safety through both offline and online methods. The online approach involves training a classifier as a run-time failure monitor to detect closed-loop, system-level failures, subsequently triggering a fallback controller that robustly handles these detected failures to preserve system safety. For the offline approach, we improve the original controller via incremental training using a carefully augmented failure dataset, resulting in a more robust controller that is resistant to the known failure modes. In either approach, the system is safeguarded against shortcomings that transcend the vision-based controller and pertain to the closed-loop safety of the overall system. We validate the proposed approaches on an autonomous aircraft taxiing task that involves using a vision-based controller to guide the aircraft towards the centerline of the runway. Our results show the efficacy of the proposed algorithms in identifying and handling system-level failures, outperforming methods that rely on controller prediction error or uncertainty quantification for identifying system failures.

\noindent Website: {\tt\footnotesize \url{vatsuak.github.io/visual-failure-mitigation}}
\end{abstract}

\section{Introduction}
\label{sec:intro}
Due to recent advancements in neural networks and deep learning, vision-based controllers (VBCs) are becoming increasingly common in autonomous systems.
These controllers enable systems to process visual data, such as images and point clouds, in real-time, facilitating significant improvements in the perception and decision-making capabilities of the system.
However, despite their benefits, VBCs are prone to errors, especially when the system operates in unseen or novel environments. 
These errors can cascade to catastrophic system-level failures, limiting their use in safety-critical applications. 
Finding such failures of VBCs and developing mechanisms to safeguard against them is the primary focus of this paper.

A key characteristic of VBCs is their susceptibility to semantic failures and environmental conditions.
For instance, in autonomous vehicles, adverse weather conditions such as heavy fog or changes in lighting can drastically reduce the controller's accuracy, leading to misclassifications or incorrect decisions.
Several methods have been developed to detect such failures, particularly through adversarial perturbations -- small, deliberate modifications to the visual input that lead to large prediction errors \cite{huang2017safety, pei2017deepxplore}. 
While these methods are effective in identifying component-level issues, they often overlook the impact of such failures on the
overall system.
Indeed, not all perception errors translate to meaningful system-level failures.
For instance, a vision module error may be inconsequential for a stationary robot but could cascade into catastrophic outcomes for a high-speed drone.
Thus, a system-level perspective on visual failures is essential.

Forward simulation-based approaches attempt to address this gap 
using system rollouts or exhaustive searches to identify counterexamples that violate system safety specifications \cite{indaheng2021scenario},\cite{fremont2020formal}.
However, in the context of VBCs, these approaches can be computationally prohibitive due to these
controllers’ high-dimensional and complicated input spaces.
In addition, the failure modes must be identified in relation to different environmental conditions in which the system might operate, which for an autonomous aircraft can include different runways, weather, or lighting conditions. This further increases the computational burden on forward simulation-based approaches to expose the system failures.

In this work, we propose a novel approach to detect and mitigate system-level failures in autonomous systems equipped with VBCs.
Our key idea is to compute the Neural Reachable Tubes (NRTs) of the closed-loop system under VBC, which are parameterized reachability-based constructs that predict the set of unsafe configurations the system may encounter under different environmental conditions. 
NRTs serve as a neural network-based counterpart to traditional Backward Reachable Tubes (BRTs) but with two key distinctions.
First, NRTs are computed via learning-based reachability methods, leveraging only the sample data from VBC. 
This allows us to overcome the computational bottlenecks of traditional reachability techniques and their reliance on analytical models that are difficult to derive for systems under VBC  due to the complex relationship between system states and visual observations.
Second, NRTs are parameterized by different environmental conditions, such as lighting and cloud coverage, providing a more diverse set of system-level failures.

Once these failures are identified, they are used to design both online and offline safety mechanisms for the system. 
For runtime safety, we use the visual failure inputs to train a simple classifier that acts as a failure detector (FD) at runtime. 
The FD is then used to trigger a fallback safety mechanism whenever a failure is detected. 
Using conformal prediction, we provide coverage guarantees on the predictive performance of FD, ensuring system safety with the overall pipeline. 

To further enhance the performance of VBC, the identified failure scenarios are used in an offline incremental retraining of the VBC. 
By retraining the VBC specifically on failure scenarios, the updated controller becomes more resilient and better equipped to handle previously encountered failure modes, resulting in an overall performance improvement.
We compare our approach against several baselines, highlighting
the effectiveness of the proposed algorithms in identifying and mitigating system-level failures of VBCs.

In summary, the key contributions of this paper are: 
\begin{itemize}
    \item we propose a framework to compute NRTs for closed-loop robotic systems under VBCs. These NRTs expose the system-level failures of VBCs across various environmental conditions;
    \item we design a runtime monitor to detect the closed-loop failures of VBCs while providing theoretical guarantees on the predictive performance of the monitor. We couple it with a fallback controller to enhance safety;
    \item we propose a targeted retraining approach for VBCs to improve their overall performance and robustness to identified failures;
    \item we demonstrate the proposed algorithms on an autonomous aircraft taxiing system that leverages an RGB image-based controller to steer the aircraft on the runway. We provide a comparison with multiple baselines, including prediction error-based and uncertainty-based failure detectors, and highlight the importance of safety from a system-level view.
\end{itemize}

\noindent \textbf{{Differences from the conference version.}} This paper significantly extends our previous work presented in \cite{10611397} with the following key advancements:
\begin{itemize}
\item \textit{Neural Reachable Tubes:} Unlike the previous work, which relied on grid-based computations for failure mining, we leverage DeepReach \cite{bansal2021deepreach} to compute NRTs. This is the first demonstration of applying NRTs for failure mining of VBCs, enabling our system to scale efficiently to various environmental conditions, such as different times of day and weather conditions, without suffering from the curse of dimensionality.

\item \textit{Comprehensive Analysis Over Environmental Conditions:} We extend our failure analysis to encompass multiple environmental conditions using a unified NRT. This allows us to systematically identify failure modes across diverse environments, providing a robust signal for training the failure detector (FD) and enhancing the system's safety.

\item \textit{Online and Offline Failure Mitigation Strategies:} We significantly advance the reliability of online FD by providing theoretical guarantees on its performance.
Specifically, we present the proof of coverage guarantees over the recall metric of the FD, ensuring that the system prioritizes safety by conservatively identifying failures. 
In addition, we propose a new offline mechanism that is based on targeted incremental training of the VBC using identified failures for mitigating the safety risks and improving its performance over time.

\end{itemize}

\section{Related Work}
\label{sec:related}
\noindent\textbf{{Failure mining of \VBC s.}} Determining failure cases in \VBC s is key to improving their safety. \textit{Verification} methods \cite{tjeng2017evaluating,katz2017reluplex,pmlr-v151-brown22b} rigorously identify regions of the input space that lead to incorrect outputs, a process known as \textit{input-output verification}. Such methods are frequently used to find failure cases of VBCs that employ neural networks (NNs) based components. However, these methods are often restricted to small models with specific components, such as Tanh or ReLU activations.

For \VBC s incorporating deep neural networks (DNNs) for processing 
image based inputs, challenges emerge due to the high dimensionality of these images. Addressing these issues often requires advanced techniques, such as generating adversarial perturbations \cite{huang2017safety} or analyzing neuron activations \cite{pei2017deepxplore}. However, these methods typically focus on isolated perception failures. In robotics, it is essential to evaluate perception modules alongside other components, including decision-making and planning, a process referred to as closed-loop or system-level verification.
 Owning to the complexity of closed loop components, one typically constructs some form of reduced representation of the original system over which the failure analysis can be tractably performed. Recent strategies have employed Generative Adversarial Networks \cite{katz2021verification}, piecewise affine abstractions \cite{hsieh2021verifying}, geometric sensor mappings \cite{santa2022nnlander}, canonical model decomposition \cite{waite2023data}, and perception contracts \cite{astorga2023perception} for obtaining such representative models. While these methods offer promising directions, such models often involve simplifying assumptions, whereas the original system can be complex and non-trivial, leading to a possible mismatch between the two. 

With the advancement of photorealistic simulators\cite{xplane, matterport}, \textit{falsification} has emerged as an important technique for system analysis \cite{xplane, matterport, indaheng2021scenario, fremont2020formal, DreossiVerifAI}. Falsification methods identify failure cases by simulating system rollouts. However, these simulations can be computationally demanding and can potentially overlook rare, long-tailed failures, particularly in high-dimensional input spaces. 

In \cite{chakraborty2023discovering}, Hamilton-Jacobi reachability was applied for falsifying \VBC s. By computing BRTs using photorealistic simulators, their method systematically identified long-tailed visual failure modes in various robotic systems. However, their grid-based approach, prone to the curse of dimensionality, limited their analysis to a small set of environmental parameters and thereby restricted the diversity of discovered failure modes. In this work, we introduce NRTs, a data-driven alternative that replicates the success of grid-based methods. The complexity of NRTs do suffer from the dimensionality issue allowing us to extend the reachability analysis across a wider range of environmental conditions.

\noindent\textbf{{Runtime Monitors for robotic systems.}} Failure analysis for vision-based controllers (VBCs) aids in preemptively determining if the system is at risk of safety violations during runtime. \textit{Runtime monitoring} \cite{RahmanCorkeEtAl2021} is a valuable tool for assessing the reliability and performance of perception-based systems during deployment, especially in unknown or unpredictable environments. Ensemble methods \cite{lakshminarayanan2017simple} have proven effective for quantifying predictive uncertainty in DNNs, enabling autonomous systems to self-monitor and self-surveil \cite{peng2023sotif}. A widely adopted approach within these methods involves generating training data by injecting artificial noise into ground truth data (e.g., images), creating a distribution shift \cite{sun2021complementing}. Learned detectors can then identify these shifts during runtime. Additional strategies for online distribution shift detection include sensor redundancy, facilitated by integrity checks \cite{balakrishnan2020integrity}. Temporal diagnostic graphs \cite{antonante2023monitoring} also enable real-time fault detection, providing formal guarantees on the maximum number of uniquely identifiable faults. In this work, we propose a neural network-based failure detection approach to identify system-level failures in autonomous systems.

\noindent\textbf{{Failure Mitigation.}} Besides anticipating failures, a robotic system should be capable of making corrective actions when deployed in \textit{a priori} unseen environments. These approaches could be broadly categorized into \textit{online} and \textit{offline} methods. Online methods are often seen to employ \textit{safety filters} to dynamically adjust the system's behavior based on the risk of failure or the likelihood of entering unsafe regions. Various methods use Control Barrier Functions \cite{ames2019control} or Hamilton-Jacobi reachability \cite{wabersich2023data} to determine when intervention is necessary. Such strategies are analogous to using a nominal controller while the system operates within the in-distribution region and switching to a \textit{fallback controller} when the safe operation is uncertain. Fallback controllers can rely on expert information from human intervention \cite{pmlr-v15-ross11a} or consist of failsafe mechanisms \cite{saxena2017learning, lin2020reachflow} designed to maintain safety with minimal external input.

Offline methods focus on developing learning algorithms that are \textit{distributionally robust}, under the worst-case scenario within a defined envelope of distributional shifts to ensure reliable OOD behavior \cite{Ben-TalHertogEtAl2013, DuchiNamkoong2021}. Another strategy is domain randomization during training \cite{tobin2017domain}, which prepares the system to handle a wide variety of potential environments. Additionally, some approaches involve provably safe controller synthesis \cite{calinescu2022discrete} for deep NN-based perception systems.

In this work, we present a simple fallback controller triggered by a learned FD to enhance system safety, demonstrating an approach to online failure mitigation. 
Finally, we propose an incremental training scheme \cite{istrate2018incremental}, which aims to improve the robustness of a trained NN using carefully selected samples from the network's OOD regions as an alternative offline route to the safety problem.

\section{Problem Setup}
\label{sec:problem}
Consider a robot in an environment, $\env$. The environment can be broadly considered to be all factors that are external to the robot  (e.g., geographical location where the robot is navigating, visual guides such as road/lane markings, or even characteristics such as different weather conditions and time of the day). 

For the purposes of this paper, we assume that these environmental properties can be characterized by a parameter $d$. For example, various aspects of a scene, such as a morning setting at an airport runway, which can include the runway lights, shadows cast by the sunlight, lane markings visibility, etc.,  can be collectively assigned to a specific value of $d$.

We model the internal factors of the robot as a dynamical system with state $\state\, \in \mathbb{R}^n$, control $\ctrl\, \in \mathcal{U}$(a compact set), and dynamics, $\dot{\state} = \dyn(\state, \ctrl)$. The robot possesses a vision-based, output-feedback policy $\policy$ (also referred to as \VBC), which takes in sensor measurements $\image$ at any state $\state$ and returns the control action $\ctrl \coloneqq \policy(\image)$. Such a VBC $\policy$  often involves NNs such as an end-to-end learning model or can be composed of different sub-modules viz decision making, planning, etc. 

The robot is equipped with a sensor $\sensor:\state \times \envp \rightarrow \image$ to obtain the measurements or observations from the environment. One can think of $\sensor$ being a camera or LiDAR and the measurement $\image$ being an image or point clouds, respectively. In this work, we specifically focus on vision sensors for which $I$ is often high-dimensional. 

We assume access to a simulator that allows us to obtain samples of $\image$ at different robot states and environmental conditions. This assumption allows us to leverage the advances in photorealistic simulators to analyze the safety of robotic systems with rich sensor inputs before their actual deployment.

Let $\traj{\policy}{\state}(\tau)$ be the robot's state achieved at time $\tau$ when it starts from state $\state$ at time $0$, and follows policy $\policy$ over $[0,\tau]$. Finally, let $\obs$ denote a set of failure states for the system. 
As an example, $\obs$ could represent obstacles for a ground robot or off-runway positions for an autonomous aircraft. Thus, an initial state $\state$ is considered unsafe for the system if $\exists s\in [0,\tau],\,  \traj{\policy}{\state}(s) \in \obs$. 
The set of observations the system sees, starting from such unsafe states, are thus considered \textit{failure inputs for the closed-loop system} (as they eventually steer the system to $\obs$) and denoted as $\ibrt$.

\noindent \textbf{Objective 1:  Mining Failures of a Vision-Based Controller.} Our first objective is to find this set of observations $\ibrt$. We term this problem as the task of \textit{mining} for the closed-loop visual failures of the system. The presence of such failure observations is not only influenced by the system's internal properties (such as the dynamics)  but also highly dependent on the external environment (such as the illumination of different objects). We hypothesize that by obtaining and analyzing such an $\ibrt$, we will uncover specific properties of the visual inputs that cause the robotic system to fail. 

Next, using the information from these mined failure inputs, we wish to formulate preemptive measures to ensure system safety during runtime. 
This can be done by introducing additional modules that take over the system's control in the face of such failure inputs in an online setting or via direct modification of the original system in an offline setup such that it is robust against inputs similar to the mined failures. 

\noindent \textbf{Objective 2: \emph{Online} Failure Detection and Fallback Mechanism.}
For the online case, we seek an additional algorithm to determine whether a given input is a failure, i.e., a failure detector (FD). Here, we are interested in obtaining a mapping, $FD: \image \rightarrow \ \{0,1\}$, that provides a binary decision of whether a given input $\image$ can possibly lead to the failure of the system: 

where $1$ means that $\image$ is failure, and $0$ means it is not. Hence, an ideal FD should output $1$ whenever $\image \in \ibrt$ and $0$ otherwise. In addition to the FD, we must design a \textit{fallback controller}. Such a controller should be able to obtain feedback from the FD and determine how the input should be processed to prevent system failure.

\noindent \textbf{Objective 3: \emph{Offline} Improvement via Controller Refinement.} For the offline version, our goal is to find a refined policy $\policy_{refined}$ that is more robust to the failures in $\ibrt$. A key consideration here is to make sure that $\policy_{refined}$ does not lose on the performance for which it was initially designed. Finally, it is important to recognize that such changes might introduce new failure modes that differ from those in the original model. Therefore, the process must be iterative, allowing for continuous refinement and modification of the system until it meets the desired performance and safety standards. 

\begin{figure}[ht]
\centering
\includegraphics[width=\columnwidth]{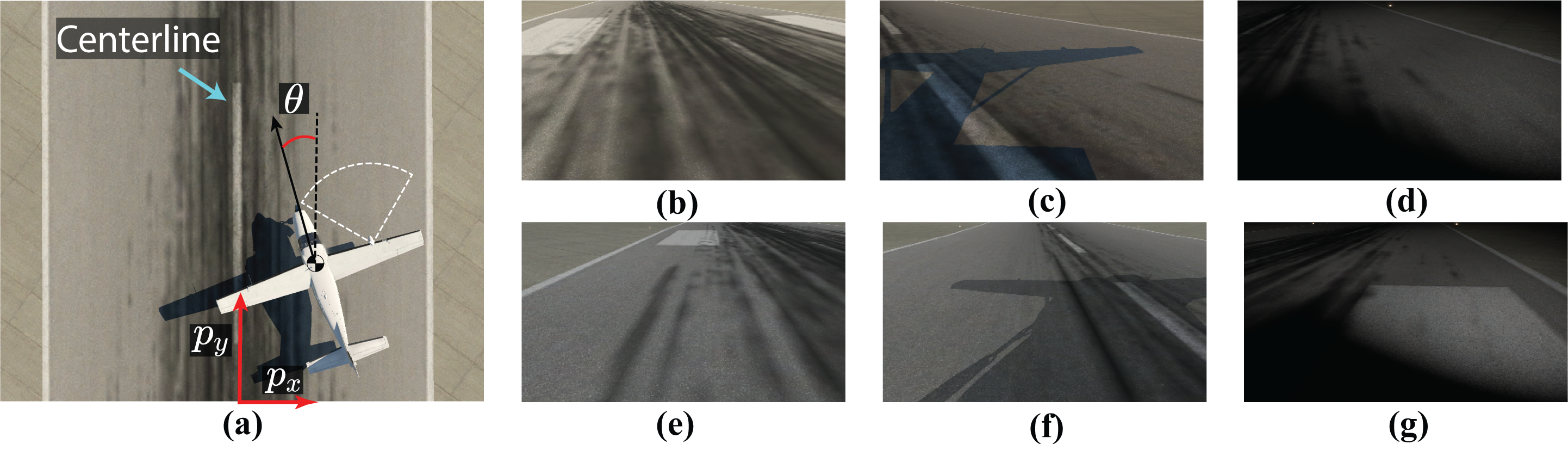}
\vspace{-0.5em}
\caption{\small{\textbf{(a)} The overhead view of the XPlane simulator with the aircraft on the KMWH runway. $p_{x}$, $p_{y}$, $\theta$ denote the state of the aircraft; dashed-white lines show FoV of the camera. The aircraft is required to track the centerline and perform the taxiing task without leaving the runway. Runway simulation images $d_2:$ \texttt{clear} and $d_1:$\textbf{(b)} \texttt{morning}, \textbf{(c)} \texttt{evening}, and \textbf{(d)} \texttt{night}, and $d_2:$ \texttt{overcast} and $d_1:$\textbf{(e)} \texttt{morning}, \textbf{(f)} \texttt{evening}, and \textbf{(g)} \texttt{night}.}}

\label{fig:taxinet_running}
\end{figure}

\vspace{0.2em}
\noindent \textbf{\textit{Running example.} TaxiNet.} We introduce the aircraft taxiing problem \cite{katz2021verification} as a running example to illustrate key concepts.
Here, the robot is a Cessna 208B Grand Caravan modeled as a three-dimensional non-linear system with dynamics:
\vspace{-0.6em}
\begin{equation}
\label{eqn:dyn_taxinet}
     f(\state,\ctrl) = [\dot p_x\quad \dot p_y\quad  \dot \theta]^T= [v\ sin(\theta)\quad v\, cos(\theta)\quad  u]^T
     \vspace{-0.6em}
\end{equation}
where state $\state$ is composed of $p_x$, the cross-track error (CTE), $p_y$, the downtrack position (DTP), and $\theta$, the heading error (HE) of the aircraft in degrees from the centreline (Fig. \ref{fig:taxinet_running}(a) shows how these quantities are measured). $v$ is the linear velocity of the aircraft kept constant at 5 m/s, and the control $u$ is the angular velocity.

The image observations are obtained using the X-Plane flight simulator that can render the RGB image, $I$, from a virtual camera ($\sensor$) mounted on the right wing of the aircraft. $\image$ depends on the state $\state$ as well as two environmental parameters: $\envp_1$: Time of day, $\envp_2$: Cloud conditions. In this work, we vary $\envp_1$ and $\envp_2$ as discrete parameters where $\envp_1$ can take 3 values: \texttt{morning}, \texttt{evening}, and \texttt{night}, and $\envp_2$ can take 2 values: \texttt{clear} and \texttt{overcast}. Depending on the parameters, we can expect a different visual input $I$ at the same $\state$ (see Fig. \ref{fig:taxinet_running}(b)-(g) for sample images). 

The goal of the aircraft is to follow the centreline as closely as possible using the images $\image$. 
For this purpose, the aircraft uses a Convolutional Neural Network (CNN), which returns the estimated  CTE, $\hat p_x$, and the estimated HE, $\hat \theta$.
A proportional controller (P-Controller) then takes these predicted tracking errors to return the control input as follows:
\vspace{-3.0em}

\begin{equation}
    (\hat p_x, \hat \theta ) = \text{CNN}(\image) \quad \ctrl \coloneqq tan(-0.74\hat p_x -0.44\hat \theta)
    \label{eqn:pctrl}
    \vspace{-0.6em}
\end{equation}

Hence, TaxiNet is essentially the policy $\policy$, a composition of the CNN and the P-Controller.
Readers are recommended to see \cite{katz2021verification} for the implementation details of the CNN that we use for this task.

We define the unsafe states for the aircraft as $\mathcal{O} = \{\state: |p_x| \geq B\}$, where $B$ is the runway width. Thus, $\obs$ corresponds to aircraft leaving the runway.
As a concrete example, for one of our simulation environments, runway 04 of Grant County International Airport (codenamed KMWH), $B = 10$. Hence, $\mathcal{O}_{\text{KMWH}} = \{\state: |p_x| \geq 10\}$. 

Our goal is to find the set of input images that drive the aircraft off the runway under the control policy in \eqref{eqn:pctrl} and eventually detect and mitigate such failures.

It is important to note that in our analysis, we evaluate the CNN and policy \emph{without modification} for objective 1. For objective 2, we introduce a fallback mechanism to demonstrate online failure mitigation, and for objective 3, we aim to improve the policy through an offline approach, while \emph{keeping the CNN structure unchanged}.

\section{Background}
\label{sec:background}
\subsection{Hamilton-Jacobi Reachability Analysis}

In this work, we will use Hamilton-Jacobi (HJ) Reachability analysis to mine the failure modes of a \VBC.
We now provide a brief overview of HJ reachability and refer the readers to \cite{bansal2017hamilton} for more details.

In reachability analysis, we focus on calculating the \textit{Backward Reachable Tube (BRT)} of the system. 
The BRT refers to the set of initial states from which an agent, starting from these states and following a state-feedback policy $\policy(\state)$, can reach the failure set $\obs$ within the time interval $[t, T]$:

\vspace{-0.6em}
\begin{equation}
\vspace{-0.6em}
\mathcal{V} \coloneqq \{\state: \exists \tau \in [t, T], \zeta^{\policy}_{\state}(\tau) \in \mathcal{O} \}
\label{eqn:brt}
\end{equation}

HJ reachability analysis allows us to compute the BRT for general nonlinear systems, even when dealing with control and disturbance inputs affecting the system within arbitrarily shaped failure sets.
Specifically, to compute the BRT, the failure set is first represented as a sub-zero level set of a function $l(\state)$, denoted as $\obs = \{\state: l(\state) \leq 0\}$ \cite{mitchell2005time,mitchell2002level}. 
The function $l(\state)$ typically represents the signed distance from a state to the failure set $\obs$. With this formulation, the BRT computation can be reframed as an optimal control problem that involves finding a value function defined as:

\vspace{-1.0em}
\begin{equation}
V(\state,t) = \min_{\tau \in [t, T]} l(\zeta^{\policy}_{\state}(\tau))
\vspace{-0.6em}
\label{eqn:vfn}
\end{equation}

This value function (defined in \eqref{eqn:vfn}) can be iteratively computed using dynamic programming principles leading to a partial differential equation known as the Hamilton-Jacobi-Bellman Variational Inequality (HJB-VI) \cite{bansal2017hamilton}:
\vspace{-1.0em}

\begin{equation}
\begin{aligned}
\min\{D_tV(\state,t)+H(\state,t)&,l(\state) - V(\state,t)\}=0 \\
\text{with }V(\state,T) = l(\state)
\label{eqn:hjivi}
\vspace{-0.75em}
\end{aligned}
\end{equation}

here $D_t$ represents the temporal gradients of the value function. The Hamiltonian, denoted as $H \coloneqq \inprod{\nabla_{\state} V(\state,t)}{f(\state, \policy(\state)}$\footnote{$\nabla_{\state} $ denotes spatial gradients and $\inprod{.}{.}$ denotes the standard inner product}, embeds the system dynamics in the HJB-VI. Essentially, \eqref{eqn:hjivi} is a continuous-time counterpart to the Bellman equation in discrete time. Once the value function is determined, the BRT is obtained as the set of states from which entry into the failure set is unavoidable. Consequently, the BRT corresponds to the sub-zero level set of the value function:

\vspace{-1.0em}
\begin{equation}
\mathcal{V} = \{\state: V(\state,t) \leq 0\}
\vspace{-0.5em}
\end{equation}

\subsection{DeepReach}
Traditional methods compute a numerical solution of the HJB-VI over a statespace grid \cite{mitchell2007toolbox}; however, for high dimensional systems, the process becomes computationally intensive, resulting in an exponential scaling of memory and computational complexity (in the systems' dimensions) often called the ``curse of dimensionality".
Instead, we use DeepReach \cite{bansal2021deepreach} to compute the value function.
Rather than solving the HJB-VI over a grid, DeepReach represents the value function as a sinusoidal NN and learns a parameterized approximation of the value function. 
Thus, the memory and complexity requirements for training scales with the value function complexity rather than the grid resolution or the system dimension. 
To train the NN, DeepReach uses self-supervision on the HJB-VI itself. 
Ultimately, it takes as input a state $\state$ and time $\tau$, and it outputs the value $V_{\theta}(\state, \envp, \tau)$, where $\theta$ are the NN parameters.
We will go into specific network details in Sec. \ref{sec:mining_failures} and refer interested readers to \cite{bansal2021deepreach} for further information about DeepReach.

\section{Mining Failures of a Vision-Based Controller}
\label{sec:mining_failures}
We cast the problem of finding closed-loop visual failures as a reachability problem. By doing so, we can apply the tools from the literature of HJ Reachability analysis to determine the specific visual inputs that drive the system to failure. We term this task as mining the failures of the \VBC. 

The key approach for this process is to obtain the reachable tube of the system we wish to analyze. The tube will then expose the specific system configurations from where the system will inevitably enter the failure region. These configurations are precisely the failure modes of the \VBC.

\subsection{Neural Reachable Tube (\NRT) Computation of Systems with Closed-Loop \VBC s}

The first step towards BRT computation is obtaining a closed-loop policy, $\hat\policy$, for the system, by composing the sensor mapping with the \VBC\, as follows,

\vspace{-0.65em}
\begin{equation}
    u = \policy(\image)\;=\;\policy(\sensor(\state, \envp))\;
    \implies \ctrl = \hat\policy(\state, \envp)\quad
    \vspace{-0.65em}
\label{eqn:ctrl_state}
\end{equation}

Given the policy $\hat\policy$, we can compute the BRT $\brt$ by solving the HJB-VI in \eqref{eqn:hjivi}.
Using $\brt$ we can obtain failure inputs as,

\vspace{-0.5em}
\begin{equation}
\vspace{-0.7em}
    \ibrt =\{I|\, I = S(\mathbf{x},\envp), \mathbf{x} \in \mathcal{V}\},
    \label{eqn:problem}
    \vspace{0.2em}
\end{equation} 

which can be subsequently analyzed to find the typical characteristics of the failure inputs.
Finding $\ibrt$ through a BRT allows us to tractably search for failures over high-dimensional input space by converting it into a search over the statespace, which is typically much lower dimensional.

However, existing HJ reachability methods typically require an analytical expression of $\hat\policy$ to solve the HJB-VI, which in turn requires an analytical model of the sensor mapping $S$. 
Unfortunately, obtaining such models of $\sensor$ remains a challenging problem, especially for visual sensors; consequently, obtaining closed-loop policy $\hat\policy$ in an analytical form is difficult. To address this challenge, we leverage an approximation technique introduced in previous work \cite{chakraborty2023discovering}. Specifically, we employ photorealistic simulators to compute the controller output across various statespace points. These simulators allow us to query $I$ and subsequently $\ctrl$ at any given state $\state$ and environment $\envp$.

Utilizing samples from photorealistic simulators for $\sensor$, we aim to obtain a parameterized BRT that accounts for various environmental conditions, such as changes in the time of the day and different cloud cover scenarios. However, these additional parameters increase the dimensionality of the system's statespace, leading to an exponential rise in the complexity of BRT computation for grid-based systems. Consequently, this complexity inhibits the use of level set methods \cite{mitchell2007toolbox}, which were previously utilized \cite{chakraborty2023discovering} to compute a numerical approximation of the BRT. 
Instead, we now obtain a data-driven parameterized value function estimator, $\dr (\state,\env,t)$, with the help of DeepReach and compute the system's Neural Reachable Tube (\NRT) as the sub-zero level set of $\dr$. $\dr$ is commonly modeled with a DNN with parameters $\theta$ whose loss function is given by, 

\vspace{-1.5em}

\begin{align*}
    \text{Loss}(\state,\envp,\tau) &\coloneq \mathbb{L}_\text{ham} (\state, \envp, \tau) + \lambda \mathbb{L}_\text{init} (\state, \envp)\\
    \mathbb{L}_\text{ham} (\state, \envp, \tau) &\coloneq \min\{D_t\dr(\state,\envp, \tau) +H(\state,\envp,\tau),\\ 
     & \qquad \qquad \qquad \qquad l(\state) - \dr(\state, \envp, \tau)\}\\
    \mathbb{L}_\text{init} (\state, \envp) &\coloneq |\dr (\state,\envp,T) - l(x)| 
\end{align*}

where $\mathbb{L}_\text{init}$ denotes the loss due to the initial condition, and $\mathbb{L}_\text{ham}$ denotes the loss emerging from satisfying the HJB-VI (see eqn. \eqref{eqn:hjivi}). The hamiltonian is hence written as $H(\state,\envp, \tau) = \nabla_{\state} \dr(\state,\envp,\tau) \cdot f(x, \hat\policy(\state,\envp))$. $\lambda$ is a hyperparameter, and $T$ is the final simulation time till we are interested in analyzing the system's safety. The input to the network is the state $\state$, the environment parameter $\envp$, and time $\tau$. Hence, the network can be trained in a self-supervised manner to obtain the value function. The \NRT\ is then obtained as,
\vspace{-0.5em}
\begin{equation}
    \NRT = \{(\state, \envp) | \dr (\state,\envp,T) \leq 0\}
\vspace{-0.5em}
\end{equation}
Thus, blending simulator-based sampling with DeepReach allows us to compute the \NRT\ over a variety of states and environment conditions from the samples of $\sensor$, in place of a rigorous model of $S$. The \NRT\ can then be used to find the failure inputs as per eqn. \eqref{eqn:problem}, as a replacement of the BRT $\brt$.

\noindent \textbf{\textit{Running example.}}
To compute \NRT s using DeepReach, we sampled a uniformly distributed set of points over the system's statespace defined over  $p_x \in [-X,X]m$, $p_y \in [100,250]m$ and $\theta \in [-30\degree, 30\degree]$ ($X$ depends on the runway, e.g., for KMWH, X = 15) and the environmental states $\envp_1$ = \texttt{morning, evening, night} and $\envp_2$ = \texttt{clear, overcast}. Using the X-Plane simulator, we collected the rendered image $\image$ at each state and environmental conditions. For each $\image$, we used the CNN to obtain the estimated state $\hat p_x \text{ and } \hat \theta$ and subsequently the closed-loop policy, $\hat\policy$ according to eqn. \eqref{eqn:pctrl}. 

The sampled states and the corresponding controllers were then used to train the DeepReach model. The model is a 3-layer feed-forward NN with 512 neurons per layer and sinusoidal activation functions. We train 5 different DeepReach models, corresponding to each of the 5 runways we wish to analyze, for $T = 10s$. (We also experimented with training a single $\dr$ model for all the runways; however, that approach led to unstable optimization due to the vastly disparate features exhibited by the different runways.)
For a specific runway, given the environmental parameters, we can obtain the value of any input ($\state$, $\envp$). A negative value implies that the system has encountered at least one image along its trajectory from the given input, which leads to failure. Thus, analyzing the image traces that the system has observed over such a failure trajectory will expose the failure modes of the \VBC . We will now review some exciting failure cases identified through our analysis on one of the runways, KMWH.

\subsection{Analyzing Failure modes of the TaxiNet \VBC\ using \NRT s}

\begin{figure}[ht!]
    \centering
    \vspace{-0.7em}
         \includegraphics[width=\columnwidth]{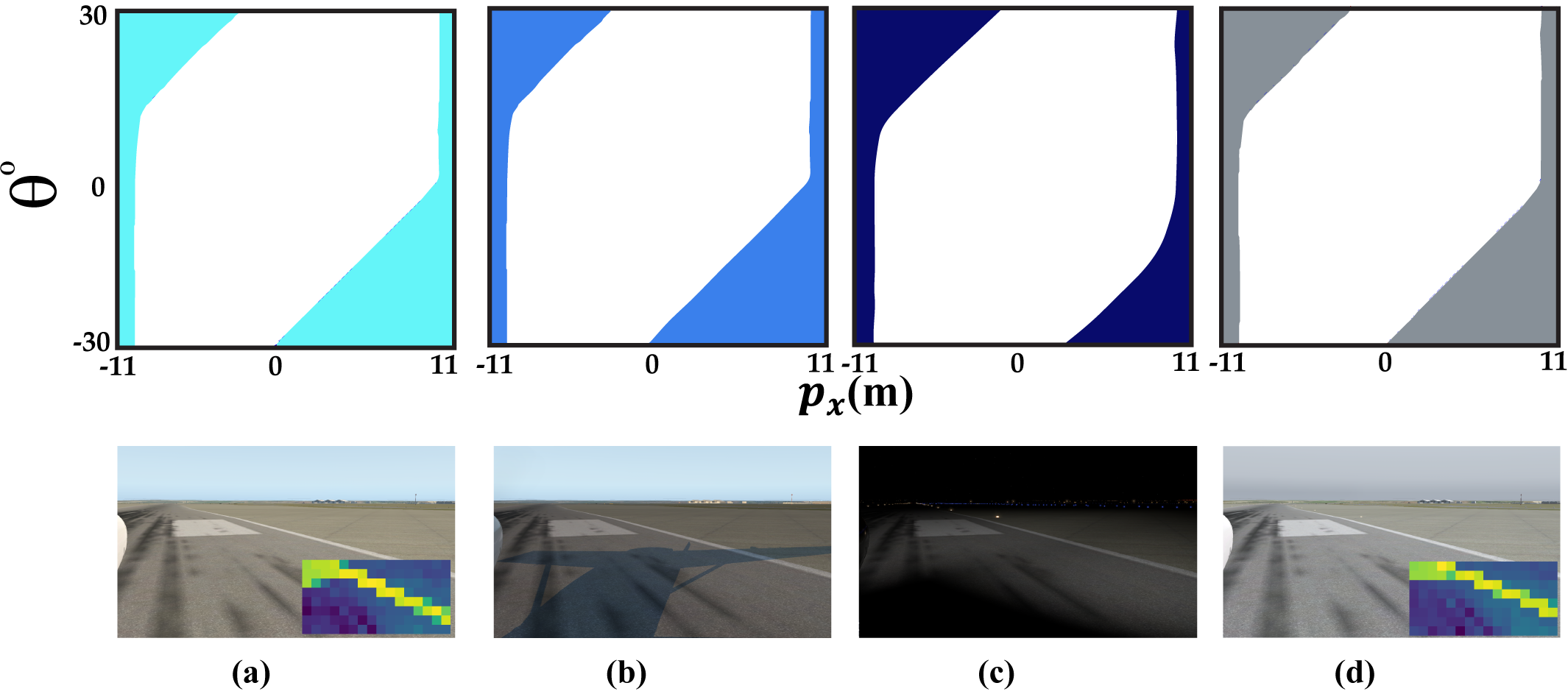}
         \vspace{-1.5em}
         \caption{Closed-loop TaxiNet \NRT\ slices over different values of $p_x$ and $\theta$ for a fixed $p_y=110m$ with variation of parameters $\envp_{i}s$ (\textbf{top row)}  and a corresponding sample image \textbf{(bottom row)} \textbf{(a)} $\envp_1 = $ \texttt{morning}, $\envp_2 = $ \texttt{clear}, \textbf{(b)} $\envp_1 = $ \texttt{evening}, $\envp_2 = $ \texttt{clear}, \textbf{(c)} $\envp_1 = $ \texttt{night}, $\envp_2 = $ \texttt{clear} and \textbf{(d)} $\envp_1 = $ \texttt{morning}, $\envp_2 = $ \texttt{overcast}, for a fixed initial starting $p_{y}=110m$. 
         Visually similar downsampled images \textbf{(bottom row inset)} for \textbf{(a)} $\envp_1 = $ \texttt{morning}, $\envp_2 = $ \texttt{clear} and \textbf{(d)} $\envp_1 = $ \texttt{morning}, $\envp_2 = $ \texttt{overcast}}
         \vspace{-0.5em}
         \label{fig:brt_tod_cc}
\end{figure}

Equipped with the ability to compute the \NRT s for the TaxiNet \VBC\, we can now analyze the different modes of failure for this controller over variations of the environmental parameters. Fig. \ref{fig:brt_tod_cc} demonstrates the 2D slices of the 5D NRT across different times of the day and cloud conditions and $p_y=110m$ for a sample runway, codenamed \textit{KMWH}. The shaded area represents the set of initial states from which the aircraft would eventually leave the runway under the VBC, whereas the white area represents the safe region.

\subsubsection{Common failure modes across parameters}

Irrespective of the values for $\envp_{i}s$, we observed that the aircraft is unsuccessful in taxiing tasks if it starts too close to the failure boundaries. The aircraft does not have enough control authority to correct its trajectory if it is too close to the failure regions ($|p_x| > 10m$); hence, from these states, the system leaves or grazes the boundary, resulting in a failure. The failure is even more pronounced when the aircraft faces the runway boundary, as indicated by the top-left and bottom-right regions of the \NRT s. In these cases, in addition to being close to the boundary, the system does not get a clear view of the center line; hence, the \VBC\ is unable to track it, resulting in a failure. We further note that the \NRT s are asymmetric about $p_x=0$ or the centreline of the runway. This is an interesting effect due to the virtual camera positioned asymmetrically on the aircraft's right wing. This allows the aircraft to have a better view of the center line from the left side, leading to better tracking abilities and a smaller BRT.
We note that these observations align with the results obtained via computing BRTs using a grid-based method \cite{chakraborty2023discovering} without losing any utility due to a learning-based approximation of the BRT. 

\subsubsection{Failure modes influenced by variations in times of day} 
It is important for any controller design to be tested against various design parameters that the system will likely encounter at runtime. Similarly, the TaxiNet controller must be effective at all times of the day. We compare its performance over 3 values for parameter $\envp_1$: \texttt{morning, evening}, and \texttt{night} (for a fixed $\envp_2 = \texttt{clear}$), by analyzing the corresponding \NRT s in Fig \ref{fig:brt_tod_cc}(a)-(c)

We note that the \NRT\ in \texttt{morning} (Fig. \ref{fig:brt_tod_cc}(a)) is similar to the one in \texttt{evening} (Fig. \ref{fig:brt_tod_cc}(b)). In \texttt{evening}, the aircraft casts a shadow on the runway, which is observed by the camera; however, we notice that it does not significantly affect the performance of the TaxiNet controller. However, at \texttt{night} (Fig. \ref{fig:brt_tod_cc}(c)), the \NRT\ is smaller than \texttt{morning} \NRT\ in the bottom-right and slightly bigger in the top-left. This is an interesting behavior, as one would expect the performance to degrade during the \texttt{night} time for a VBC, while the \NRT s suggest that TaxiNet performs better at \texttt{night}.

\begin{figure*}[ht!]
\centering
\includegraphics[width=\textwidth]{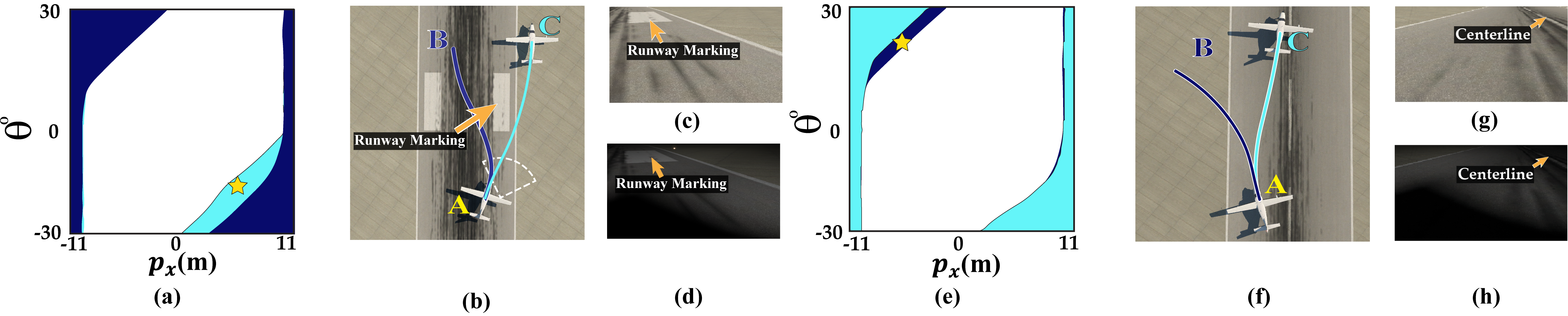}
\vspace{-1.5em}
\caption{
    \textbf{(a)} The overlaid \NRT s for $\envp_1=$\texttt{night} (blue) on $\envp_1=$\texttt{morning} (cyan) for $p_y = 110m$. The state of interest, shown with a yellow star, is only contained in the \texttt{morning} \NRT\ and not in the \texttt{night} \NRT. \textbf{(b)} Top-view of the runway in the \texttt{morning}. The trajectory, ``A" to ``C", followed by the aircraft under the CNN policy (cyan line), takes it off the runway in the \texttt{morning}. The trajectory (blue line) from ``A" to ``B" is followed at \texttt{night}.\textbf{(c)} The runway marking in the image, which acts as a failure mode, can be vividly seen by the CNN at location ``A" in the \texttt{morning} cannot be seen clearly at \texttt{night} \textbf{(d)} due to poor illumination. \textbf{(e)} The overlaid \NRT s for $\envp_1=$\texttt{morning} (cyan) on  $\envp_1=$\texttt{night} (blue) for $p_y$ = 190m. The state, shown with a yellow star, is only included in the \texttt{night} \NRT. \textbf{(f)} Top view of the runway. In the \texttt{morning}, the CNN policy accomplishes the taxiing task by taking the cyan trajectory from ``A" (yellow star in (a)) to ``C." At \texttt{night}, the policy takes the aircraft outside the runway along the blue trajectory from ``A" to ``B". \textbf{(g)} The centreline in the image can be seen clearly by the CNN at location ``A" in the \texttt{morning}, whereas it cannot be seen at \texttt{night} \textbf{(h)} due to poor illumination.}
\vspace{-1em}
\label{fig:failure_over_tod}
\end{figure*}

To understand this behavior, we simulate the aircraft trajectory from a state in the \NRT\ (marked with the yellow star in Fig. \ref{fig:failure_over_tod}(a). The aircraft trajectory is shown in cyan in Fig. \ref{fig:failure_over_tod}(b) when simulated in the \texttt{morning}. We realize that the CNN confuses a runway side strip marking with the centreline (Fig. \ref{fig:failure_over_tod}(c)) and steers the aircraft towards it, ultimately leading it into the unsafe zone ($|p_x| \geq 10m$). At \texttt{night}, however, the lower visibility prevented the VBC from detecting the misleading runway marking (Fig. \ref{fig:failure_over_tod}(d)), allowing the aircraft to complete its taxiing successfully (blue trajectory in Fig. \ref{fig:failure_over_tod}(b)). This demonstrates that \emph{images containing well-illuminated runway} markings behave as failure modes of the TaxiNet controller.

Finally, as expected, the low visibility at \texttt{night} plagues the controller in other configurations. We show such an example state in Fig. \ref{fig:failure_over_tod}(e) (marked with the yellow star). The observed images by the aircraft (Fig. \ref{fig:failure_over_tod}(g, h)) reveal that, due to poor visibility during \texttt{night}, the CNN is no longer able to clearly see the centerline, causing an error in its predictions and ultimately leading the aircraft off the runway (Fig. \ref{fig:failure_over_tod}(f)).

\subsubsection{Failure modes influenced by variations in cloud conditions} 
We now study the effect of cloud conditions on the \NRT s. Fig. \ref{fig:brt_tod_cc} shows the \NRT s for $\envp_1 =$ \texttt{morning} \textbf{(a)} $\envp_2 =$ \texttt{clear} and \textbf{(d)} $\envp_2 =$ \texttt{overcast} conditions. We note that the \NRT s are actually quite similar. 

Hence, we conclude that the controller handles the variation in cloud conditions quite well. This was not surprising since, in the internal workings of the CNN, the image goes through a pre-processing step that applies an averaging filter. This enables the controller to pick out distinguishing features, such as lane markings, while downplaying the effects of color variations (the main variations brought about by the different cloud conditions). Fig. \ref{fig:brt_tod_cc} show the RGB-image observations by the camera and the corresponding downsampled version (bottom row in insets) for the \textbf{(a)} clear and the \textbf{(d)} overcast conditions. Even though the raw RGB images show discerning properties, the downsampled version in both cases looks quite similar, leading to comparable performance of the TaxiNet over various cloud conditions.

\noindent \textbf{{Computation time comparisons.}}
 Our method took $\thicksim$15 hours to compute the \NRT.
The majority of this time was spent computing $\hat\policy$, which involves rendering images at state samples and querying the CNN (TaxiNet). We used around $\thicksim10^6$ samples of the statespace to train the DeepReach model (the training took 2 hrs). On the other hand, computing a 5D parameterized BRT using a grid-based method (such as \cite{mitchell2007toolbox}) takes a total of  $\thicksim$19 hours for the same procedure. Robotic systems often employ higher dimensional statespace; hence, we expect this gap to increase further for high-dimensional spaces.

Alternatively, instead of opting for HJ reachability-based methods, one may choose to use exhaustive search and subsequently roll out trajectories starting from all areas of importance over the system's statespace to analyze the failure modes. We estimate that it would require $\thicksim70$ days to complete such an analysis over the $10^6$ trajectories, with each trajectory taking $6s$ on average to complete; the primary bottleneck coming from a combination of \textit{repeated} image rendering (accounting for $\sim$40\% of rollout times), forward CNN calls, and post-processing of the CNN predictions for the entire duration of the trajectory simulation. On the other hand, our proposed reachability-based approach leverages dynamic programming and advances in deep learning to alleviate the need for repeated system queries and fast gradient-based optimizations.

\section{Online Failure Detection and Fallback Mechanism}
\label{sec:anomaly_detector}
In the previous section, we demonstrated how \NRT s can be used for identifying system-level failures of a \VBC .
However, the proposed analysis is primarily suitable for offline computations and cannot be used at system runtime. Hence, in this section, our objective is to design a failure detector (FD) that can learn the underlying pattern of inputs contributing to a failure. Such an FD can then be used online to flag potential failure inputs and trigger a fallback controller to ensure system safety.

\subsection{Designing a Failure Detector}
Our FD design uses a learned mapping $\classy_{\phi}: \image \rightarrow \mathbb{R}$, where $\phi$ represents the learnable parameters that characterize $\sigma$. As before, $\image$ is an observation from perception sensors, such as camera images or point clouds from LiDARs. To train $\classy_{\phi}$, we assume access to a labeled dataset $\{\image_i, Y_i\}_{i=1:N}$ of size $N$ where $Y_i \in \{0,1\}$ is the label corresponding to observation $\image_i$. If $Y_i = 1$, it denotes that $\image_i$ can cause a system-level failure, while $Y_i = 0$ denotes a safe observation. A well-trained $\classy_{\phi}$ can then be used to classify any novel observation, $\image_{\text{test}}$ as safe or unsafe according to a simple threshold $\hat q \in [0,1]$,
\begin{equation}
FD(\image_{\text{test}}) = \begin{cases}
                                1 & \quad \text { if } \classy_{\phi}(\image_{\text{test}}) \geq 1 - \hat q \\
                                0 & \quad \text { otherwise }
                            \end{cases}
\label{eqn:inference}
\end{equation}

Our key idea is to utilize the mined failure inputs to construct a failure dataset for training the $\classy_\phi$ used for the FD. We hypothesize that at runtime, the FD will implicitly leverage the revealed failure modes to assess whether a previously unseen (i.e., test) input observation has the potential to trigger a system failure.

We use the system's \NRT\ under \VBC\ to obtain the training samples as follows,
\vspace{-0.4em}
\begin{align*}
    \{\image_i, Y_i\} = \{ \sensor(\state_i,\envp_i), \mathbbm{1}((\state_i, \envp_i)\in \NRT)\} \, \text{ where } (\state_i, \envp_i) \sim U
\end{align*}

Here $\mathbbm{1}$ denotes the indicator function, and $U$ is any sampling distribution, such as the uniform distribution.
Specifically, we (a) randomly sample pairs of states and environmental conditions $(\state_i, \envp_i)$, (b) render the corresponding observations $\image_i$ via the sensor $\sensor$ (this could be either a photorealistic simulator that we used in the previous section or an explicit model), and (c) automatically label $\image_i$ as safe or failure depending on a binary decision of whether the sample $(\state_i, \envp_i)$ is outside or inside the \NRT.
Thus, using \NRT , our method automatically labels inputs that lead to system-level failures (that go beyond visual module failures), without requiring manual labeling.

\subsection{Failure-Detection Performance Guarantees} 
\label{sec:confp}

Now that we have a learned classifier based FD, we can employ it to detect failures of systems with \VBC s.
However, being a stochastically trained model, the FD can make erroneous predictions. Hence, we must provide proper certificates for the FD's performance since it is designed specifically for use in safety-critical scenarios. In this work, we leverage conformal prediction (CP) to provide coverage guarantees.

CP has proven to be a promising algorithm for distribution-free uncertainty quantification for prediction algorithms \cite{vovk2005algorithmic}. The main idea behind the approach is to convert an algorithm's predictions into a conformal set having strong coverage properties. CP allows us to construct a conformal set $\confclassy(\image_{\text{test}})$ that contains all the possible classes to which the test input $\image_{\text{test}}$ could belong while providing the following assurance,
\begin{equation}
    P(Y_{\text{test}}\in \confclassy(\image_{\text{test}})) \geq 1-\alpha
    \label{eqn:conf_pred}
\end{equation}
 $\image_{\text{test}}$ is the sample observation from our sensor, $Y_{\text{test}}$ is the ground truth label of whether $\image_{\text{test}}$ is a failure\footnote{In general, we can use any other object $X$ instead of $\image$ as long as they have their corresponding labels $Y$ and satisfy the exchangeability assumption.}, and $1-\alpha$ is a confidence level. This essentially means that the probability of the true prediction being included in the conformal set is greater than $1-\alpha$, $\alpha \in [0,1]$.

We wish to provide formal guarantees on the recall metric of our FD using CP. However, the guarantees obtained using the standard implementation of CP presented in eqn \eqref{eqn:conf_pred} provides assurances on the marginal distribution (over the entire dataset); in contrast, tackling the case of the recall metric warrants a form of conditional guarantee on the unsafe class. Hence, we look into \textit{class-conditioned conformal prediction} \cite{vovk2012conditional}, \cite{ding2024class}, which modifies the prior assurances to include a conditional property,
\begin{equation}
    P(Y_{\text{test}} \in \confclassy(\image_{\text{test}})| Y_{\text{test}}=1) \geq 1-\alpha
    \label{eqn:class_conf_pred}
\end{equation}
Implementing class-conditioned CP primarily requires an in-distribution calibration dataset. This dataset contains pairs of $\{\image, Y\}$ that are not observed during any previous training procedure, and \emph{contains only those data points whose label is the unsafe class, i.e., $Y = 1$}. This allows us to compute a decision threshold $\hat q$ and subsequently $\confclassy$ as follows,

\begin{enumerate}
    \item Define the classifier logits, $\hat{y} := \classy_{\phi}(\image)$ as the heuristic notion of uncertainty, where $\classy_{\phi}$ is the learned mapping (parameterized by $\phi$). $\hat{y}$ can be thought of as the probability (as decided by the FD) of the sensor observation $\image$ being unsafe.
    \item Define the nonconformity score, $s(\image,y):= y\cdot(1-\classy_{\phi}(\image)) + (1-y)\cdot \classy_{\phi}(\image)$. Larger scores imply higher disagreement between the label and prediction of $\classy_{\phi}$.
    \item Using the in-distribution calibration set, find the logit corresponding to the $1-\alpha$ quantile,
    $\hat{q}$:= Quantile\footnote{Quantile(data, p) returns the $p^{th} $ quantile of the data}(calibration scores, $ \frac{\lceil(n+1)(1-\alpha)\rceil}{n}$), where $n$ is the number of samples in the dataset. We obtain the calibration scores scores by applying the score function $s$ (from step 2) to the dataset elements (from step 3).
    \item Define $\confclassy(\image)$, the conformal set as,
    \begin{equation}
    \confclassy(\image) = \begin{cases}
                                    1 & \quad \text { if } \classy_{\phi}(\image) \geq  1 - \hat q \\
                                    0 & \quad \text { otherwise }
                                \end{cases}
    \label{eqn:conf_pred_classification}
    \end{equation}
    where $\hat{q}$ is the decision boundary corresponding to the desired $1-\alpha$ confidence.
\end{enumerate}

We make two important notes regarding $\confclassy$: \textbf{(1)} $\confclassy$ is a binary classifier (as defined in eqn. \eqref{eqn:conf_pred_classification}) with the same structure as our proposed failure detector (eqn. \eqref{eqn:inference}) and, \textbf{(2)}  By design, the $\confclassy$ ensures $P(Y \in \confclassy(\image)| Y = 1) \geq 1-\alpha$\footnote{This result follows from the quantile lemma [Sec 1.1 \cite{tibshirani2019conformal}] and relies on the exchangeability assumptions for the calibration dataset. For a detailed proof, see\cite{vovk2012conditional}, \cite{ding2024class}}.

The first property enables us to directly utilize $\confclassy$ as our FD, while the second allows us to leverage the guarantees of the class-conditioned CP to establish a lower bound on the recall of our FD. We formalize this second claim in the form of a simple lemma.

\begin{lemma*}\label{lemma:conf_recall}
\textit{Conformalized Recall. }Consider a conformalized classifier, $\confclassy(\image)$,
\begin{equation}
\confclassy(\image) = \begin{cases}
                                1 & \quad \text { if } \classy_{\phi}(\image) \geq  1 - \hat q \\
                                0 & \quad \text { otherwise }
                            \end{cases}
\end{equation}
that satisfies eqn. \eqref{eqn:class_conf_pred} for some $ \hat q \in [0,1]$ and a mapping $\classy_{\phi}: \image \rightarrow\mathbb{R}$,  then Recall($\confclassy$) $\geq 1-\alpha$., where,

\begin{equation*}
    Recall(\confclassy) = \dfrac{True\ Positive}{True\ Positive + False\ Negatives}
\end{equation*}

\end{lemma*}

\textit{Proof.} The key idea for this proof is to show that the definition of recall is equivalent to the coverage guarantee from eqn. \eqref{eqn:class_conf_pred} that is already satisfied by $\confclassy$.
Starting with the definition of recall, 
\begin{align*}
\label{eqn:conf_pred_recall_proof}
Recall(\confclassy) &= \dfrac{True\ Positive}{True\ Positive + False\ Negatives}\\
 &= \dfrac{True\ Positive}{Positive\ Samples}\\
 \tag*{[Assume that Y=1 are the positive samples]}\\
 &= \dfrac{P(\confclassy(\image) = 1, Y = 1)}{P(Y = 1)} \\
 &= P(\confclassy(\image) =1| Y = 1) \\
 &= P(Y = \confclassy(\image)| Y = 1)
  \tag*{[Conditional Substitution]}
 \end{align*}

Since, the conformal set $\confclassy$ is actually a binary classifier that returns a single element (and not a set), the following holds,
 \begin{equation*}
     (Y = \confclassy(\image)) \equiv (Y \in \confclassy(\image))
 \end{equation*}

 Allowing us to write,
 \begin{align*}
     Recall(\confclassy) \equiv P(Y \in \confclassy(\image)| Y = 1) \geq 1-\alpha
 \end{align*}

 The above inequality (R.H.S) is true since by construction $\confclassy$ satisfies the coverage guarantee (eqn. \eqref{eqn:class_conf_pred}). Therefore, we can conclude,

  \begin{equation}
     \boxed{RECALL(\confclassy) \geq 1-\alpha}
     \label{eqn:conform_recall}
 \end{equation}

\begin{table*}[ht]
\small
\centering
\caption{Performance comparison of learned anomaly detector on all runways in $\envp_1 =\texttt{evening}$ time with offline-calibrated on in-distribution data and online-calibrated on privileged test data conformal prediction.($(C) \implies \envp_2 = \texttt{clear}, (O) \implies \envp_2 = \texttt{overcast)}$}
\begin{tabular}{|c|cccc|cccc|cccc|}
\hline
\rule{0pt}{1\normalbaselineskip}
\multirow{3}{*}{\textbf{Airport ID}} & \multicolumn{4}{c|}{\textbf{Without CP}}                   & \multicolumn{4}{c|}{\textbf{Offline IID Calibrated CP}}    & \multicolumn{4}{c|}{\textbf{Online Re-Calibrated CP}}     \\ [1mm] \cline{2-13}  \rule{0pt}{1\normalbaselineskip}
                                     & \multicolumn{2}{c}{\textbf{Recall (\%)}} & \multicolumn{2}{c|}{\textbf{Accuracy (\%)}} & \multicolumn{2}{c}{\textbf{Recall (\%)}} & \multicolumn{2}{c|}{\textbf{Accuracy (\%)}} & \multicolumn{2}{c}{\textbf{Recall (\%)}} & \multicolumn{2}{c|}{\textbf{Accuracy (\%)}} \\ 
                                     & (C)               & (O)             & (C)                & (O)               & (C)              & (O)              & (C)                & (O)               & (C)              & (O)              & (C)                & (O)               \\ [1mm] \hline \rule{0pt}{1\normalbaselineskip}
KMWH                                 & 98.8              & 96.5            & 95.3               & 96.9              & 99.5             & 98.3             & 90.5               & 94.8              & 99.7             & 98.5             & 87.3               & 93.5              \\
KATL                                 & 99.4              & 99.2            & 97.5               & 98.8              & 99.9             & 99.8             & 94.3               & 97.9              & 99.9             & 99.9             & 92.4               & 97.2              \\
PAEI                                 & 98.1              & 97.7            & 99.1               & 99.0              & 99.2             & 98.7             & 98.8               & 98.9              & 99.5             & 99.1             & 98.5               & 98.7              \\
KEWR                                 & 95.3              & 96.7            & 94.9               & 95.0              & 98.1             & 98.8             & 89.0               & 91.1              & 98.7             & 99.3             & 84.8               & 88.6              \\
KSFO                                 & 99.4              & 99.7            & 84.8               & 82.1              & 99.9             & 99.9             & 72.3               & 70.1              & 99.9             & 99.9             & 67.0               & 64.8              \\ [1mm] \hline 
Mean                                 & \multicolumn{2}{c}{98.0}            & \multicolumn{2}{c|}{94.3}              & \multicolumn{2}{c}{99.2}            & \multicolumn{2}{c|}{89.8}              & \multicolumn{2}{c}{99.5}            & \multicolumn{2}{c|}{87.3}              \\ \hline
\end{tabular}
\label{table:comparison_table}
\end{table*}

\subsection{Online Failure Detection for the TaxiNet \VBC}
With a process for obtaining an FD with assured recall guarantees using CP, we now go into its implementation for the TaxiNet \VBC.

\vspace{0.5em}
\noindent \textbf{\textit{Failure Detector Design.}}
Our FD is the conformalized binary classifier $\confclassy$ (eqn. \ref{eqn:conf_pred_classification}), with a pre-trained EfficientNet-B0 \cite{tan2019efficientnet} as $\classy_{\phi}$, fine-tuned on the failure dataset. $\classy_{\phi}$ takes a 224x224x3 input image and returns the probability of the input being a failure image.
To generate a failure dataset, we use \NRT s computed over three different runways (codenamed: KMWH, KATL, PAEI), while we reserve 2 other runways  (codenamed: KSFO, KEWR) as test environments. To test generalization, we exclusively use the environmental parameter $\envp_1 =$ \texttt{morning} \text{ and } \texttt{night} in the training dataset and $\envp_1 =$ \texttt{evening} for the testing dataset. We sample from two cloud conditions $\envp_2 = \texttt{clear} \text{ and } \texttt{overcast}$ for both cases. This results in 12 different environmental conditions from which we sample 240K images for training and another 6 different environmental conditions from which we sample 120K images for testing.

We train the classifier using cross-entropy loss and Adam optimizer for 30 epochs with a fixed learning rate of $3e^{-4}$, which takes around 1.5 hours on an NVIDIA RTX 3090 GPU.

\begin{figure}[h]
\centering
\includegraphics[width=0.8\columnwidth]{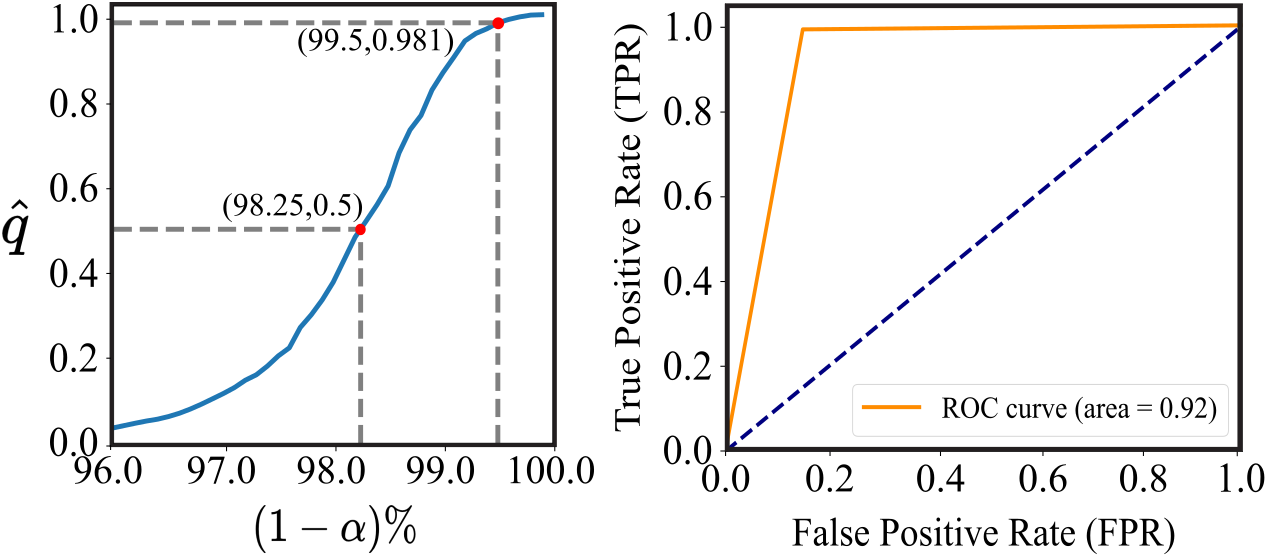}
\caption{\small{\textbf{(Left)} Variation of $\hat{q}$ on changing the values of $\alpha$. \textbf{(Right)} ROC plot on unseen test environment.}}
\vspace{-1em}
\label{fig:q_hats_and_roc}
\end{figure}

\vspace{0.5em}
\noindent \textbf{\textit{Conformalizing the Decision Threshold.}} With a trained FD, we now need to determine a threshold $\hat q$ (see eqn. \eqref{eqn:conf_pred_classification}) to accomplish our classification task. Using the class conditioned CP algorithm (Sec. \ref{sec:confp}), we obtain the values of $\hat q$ corresponding to the desired recall guarantees $(1-\alpha)$ (Fig. \ref{fig:q_hats_and_roc}(left)) using an in-distribution calibration dataset of 40K samples. We notice a steep drop in the values of $\hat{q}$ as $1 - \alpha$ goes below 0.995, which shows that (a) the classifier is extremely confident in its predictions, and (b) the classifier logits are a good measure of its confidence values. This observation is also upheld by the FD's ROC curve in Fig.  \ref{fig:q_hats_and_roc}(right), demonstrating a consistently high true positive rate.

Even using a heuristic $\hat{q} = 0.5$, a common threshold for classification problems, we can expect to achieve a recall guarantee of 98.25\% as shown in Fig. \ref{fig:q_hats_and_roc}(left), demonstrating its effective performance in safety-critical scenarios. 
However, we choose $\hat q = 0.981$ to expect a high recall guarantee of $99.5\%$ for the FD.

\vspace{0.5em}
\noindent \textbf{\textit{Evaluation.}} Table \ref{table:comparison_table} (left and middle columns) compares the performance metric of the two $\hat q$ values, (a) 0.5 and (b) 0.981, on all runways at test time $d_1 = \texttt{evening}$. We make two important observations: \textbf{(1)} The recall value using the conformalized $\hat q$ is higher than the heuristic $\hat q$, achieving a recall value of 99.5\% or higher for the training environments. and \textbf{(2)}The accuracy value using the conformalized prediction is sometimes lower than the heuristic counterpart. This is the direct result of using a higher $\hat q$ in the case of CP. All the predictions for which the network is \textit{not extremely confident} to be safe are now classified as unsafe, thereby increasing the recall values. At the same time, many of the inputs correctly classified as safe will now be incorrectly classified as unsafe due to the stringent requirement of the recall guarantee, thereby reducing the accuracy.

\begin{remark}
CP provides guarantees while assuming IID sample distribution (or, at minimum, exchangeable) for the test case, which, however, cannot be ensured for samples across different runways. Furthermore, our calibration dataset did not contain representatives from the test runways and test times of day (evening), and hence, we should not expect the guarantees from conformalized prediction to be reflected precisely in our testing cases presented in Table \ref{table:comparison_table}(middle column). A quick way to remedy this would be to add some \textit{privileged} samples from the testing scenarios in the calibration dataset. We present the performance metric for online re-calibration using privileged test samples in Table \ref{table:comparison_table}(right column). The recall values get closer to the desired 99.5\% value, demonstrating that CP can effectively provide guarantees on IID samples in new environments via online re-calibration. We will defer a detailed exploration of this direction to future work.
\end{remark}

\begin{figure*}[ht!]
\centering
\includegraphics[width=\textwidth]{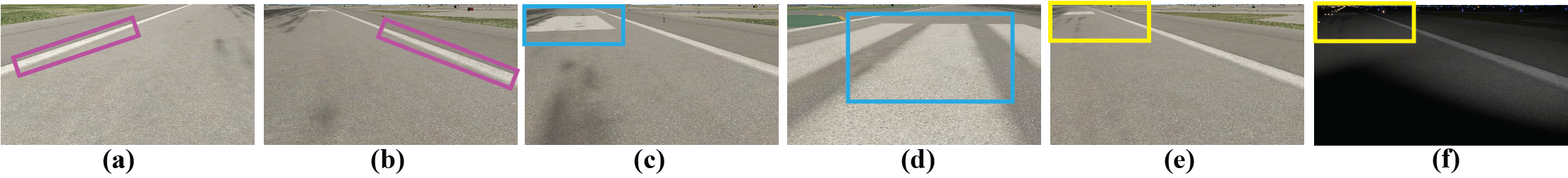}
\caption{\small{Some of the failures detected by FD. \textbf{(a, b)} Images correspond to the aircraft being close to the runway boundaries (highlighted with the purple bounding boxes).\textbf{(c, d)} TaxiNet confuses the runway markings  (highlighted with the blue bounding boxes) with the centerline and ultimately leads to a system failure. \textbf{(e, f)} Image (f) is (accurately) not classified as a failure during the \texttt{night} time (the same image is classified as a failure during the day, shown in (e)), as the runway lights (highlighted with the yellow bounding boxes) help TaxiNet predict its position accurately and thereby avoid failure.}}
\vspace{-1em}
\label{fig:all_failures}
\end{figure*}

We now showcase some of the input images that are predicted unsafe by the conformalized FD in Fig. \ref{fig:all_failures} for a qualitative analysis. We noticed that FD was able to figure out the safe limits of the runway and discern that inputs too close to the runway boundary might result in failure. Such images (white lines in Fig. \ref{fig:all_failures} (a,b)) are characterized by the closeness to the runway boundary lines where TaxiNet may be unable to estimate the aircraft's state correctly and might fail under \VBC . FD was also able to learn that images containing runway markings cause failures as the aircraft assumes them as centerline (Fig. \ref{fig:all_failures}(c,d)).

Another interesting aspect learned by the FD is that, in some runways, a similar image \textit{will not} cause a system failure during \texttt{night} time because of the runway lights, which, when lit, help TaxiNet to localize the aircraft more accurately (runway lights are at the centerline), thus avoiding a failure (Fig. \ref{fig:all_failures}(e,f)).

Such an emergent understanding of semantic failure modes is quite impressive since, in the previous sections, we had to manually analyze the \NRT s to recognize these patterns as failures. Our FD justifies using a learned classifier by practically automating the detection process without requiring any hard-coded heuristics or manual intervention.

\vspace{0.5em}
\noindent \textbf{\textit{Comparison with other FD Methods.}}
We now compare our algorithm to other baselines that are commonly used in the existing literature for the task of failure detection.

\subsubsection{Prediction Error-based Failure Detection}
For this baseline, we use a prediction error-based labeling scheme instead of the proposed HJ Reachability-based scheme to collect the training dataset for FD. This study will allow us to understand the relevance of using labels from \NRT s while training the classifier. 

Specifically, if the TaxiNet prediction error is above a certain threshold for a particular image, then we label this image as a failure. To implement this baseline, we assume that we have the ground truth state estimate for the given image during the training time, which we can get from the X-Plane simulator in our case study.
Fig. \ref{fig:all_baselines}(a) shows the prediction error-based failure labels (red) overlaid on top of the \NRT\ -based labels (blue) for one of the environments.

Evidently, the prediction error-based labels may not be a good representative of system-level failures. 
For example, the states near the green star are unsafe as per prediction error but do not actually cause the system failure (the green trajectory in Fig. \ref{fig:all_baselines}(c)), resulting in a pessimistic FD and hampering the system performance. In this case, we hypothesize that the planning module is able to handle the errors induced by the vision module along the trajectory. This is an apt example of a modular failure that only affects the CNN-based state estimator but is not detrimental to the system as a whole.

On the other hand, certain states and images may have a small error from the TaxiNet module perspective (states near the yellow star) and are not classified as failures; yet, they cascade to a system failure (the yellow trajectory in Fig. \ref{fig:all_baselines}(c)).
This results in an overly optimistic nature of the prediction error-based FD in certain regions.
This ``unpredictable'' nature of the prediction error-based FD persists with the change in the error threshold used for labeling (Fig. \ref{fig:all_baselines}(b)).
Unsurprisingly, we observe the same phenomenon in the FD trained on these prediction-error-based labels and, thus, omit those results for brevity.
This leads us to conclude that component-level, prediction error-based labels may not be a good representative for determining system-level failures.

\subsubsection{Ensemble Uncertainty-Based Failure Detection}

Another popular mechanism to detect failures is on the basis of the predictive uncertainty of an ensemble of NNs\cite{lakshminarayanan2017simple}.

\begin{figure*}[ht!]
\centering
\includegraphics[width=0.80\textwidth]{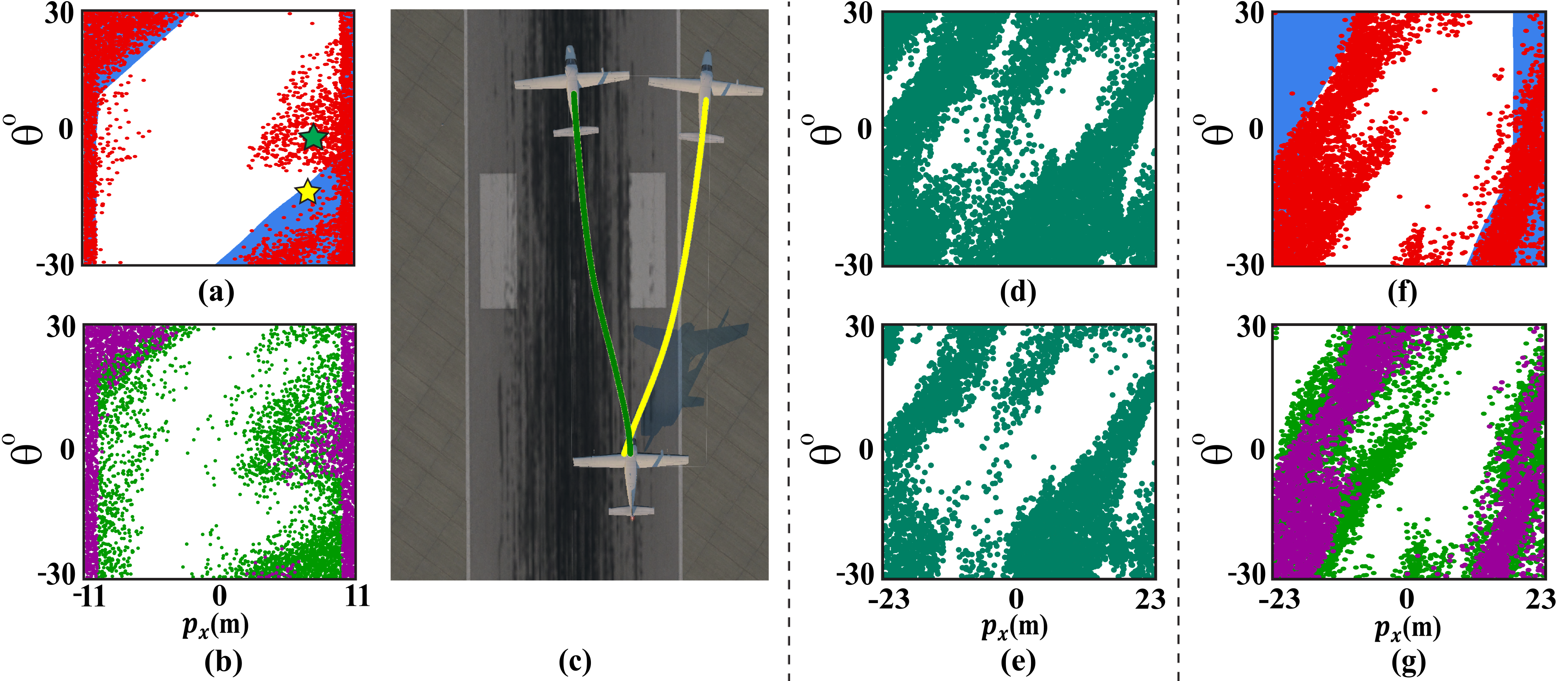}
\vspace{-0.75em}
\caption{\small{\textbf{(a)} Comparison between prediction error-based labels for $threshold=0.45$ (red) and \NRT -based labels (blue). \textbf{(b)} Prediction error-based labels for $threshold=0.3$ (green) and $threshold=0.6$ (purple). \textbf{(c)} Green and Yellow lines show trajectories starting from the green and yellow stars in (a), respectively. Labels generated using ensembling denoting failures (torquoise) and success (white) for \textbf{(d)} $threshold=0.25$, \textbf{(e)} $threshold=0.32$. \textbf{(f)} Failure labels generated using SCOD for $threshold=0.45$ (red) and \NRT\ (blue). \textbf{(g)} Failure labels generated using SCOD for $threshold=0.35$ (green) and $threshold=0.55$ (purple).}}
\vspace{-1em}
\label{fig:all_baselines}
\end{figure*}

To design an ensemble, we train 5 different TaxiNet versions with different weight initializations. If the variance between predictions exceeds a threshold for any input image, we assign that a failure. The corresponding failures over the statespace for different uncertainty thresholds are shown in Fig. \ref{fig:all_baselines}(d,e). We observe that this method does not perform well because the ensemble confidently makes incorrect predictions for some states (around the top left of statespace), leading to faulty labels in those states. On the other hand, for some states (near the central region of statespace), the ensemble disagreed on the predictions, leading to states being incorrectly marked as a failure. Finally, as seen from Fig \ref{fig:all_baselines}(d,e), the threshold choice is critical to the performance of ensemble-based FD. Therefore, it necessitates thorough analysis prior to determining a specific value.
Hence, an ensembling-based approach also fails to predict system-level failures accurately in our case.

\subsubsection{Sketching Curvature-Based Failure Detection}
Sketching Curvature for OoD Detection (SCOD) \cite{SharmaAzizanEtAl2021} presents an effective algorithm for detecting failure inputs to DNNs by computing the Fisher Information Matrix based on the weights of the trained model. This matrix characterizes how small changes in the weight space affect the DNN's probabilistic predictions. With a pre-trained DNN, matrix sketching tools can be employed to analytically compute an approximate posterior (offline). This approximation is then used at runtime to generate a meaningful failure signal or atypicality for the DNN's prediction.

We applied this method to TaxiNet to produce an uncertainty estimation for each prediction, classifying inputs with high uncertainty as failures. In essence, this method considers an input to be a failure if the gradient of the input with respect to DNN's weights is significantly orthogonal to the gradients observed during training. Fig. \ref{fig:all_baselines}(f) illustrates the detected failures in a test environment. The thresholds set the level of orthogonality allowed before classifying an input as a failure (Fig. \ref{fig:all_baselines}(g)).

While promising compared to the other two baselines, it fails to represent the true failures, particularly around the left half of the statespace. In addition, it should be noted that this method also focuses on modular failures instead of system-level failures. Even if an input is considered a failure via this method, the downstream process may be able to correct the mistakes made by the vision module and ensure system safety.

\vspace{0.5em}
\noindent \textbf{\textit{Failure Modes of FD.}}
Even with guarantees on the predictive performance of the FD, we were still unable to identify certain types of system failure modes.
Such misclassification happens particularly for the images corresponding to the states near the \NRT\ boundary. 

\begin{figure*}[ht!]
\centering
\includegraphics[width=\textwidth]{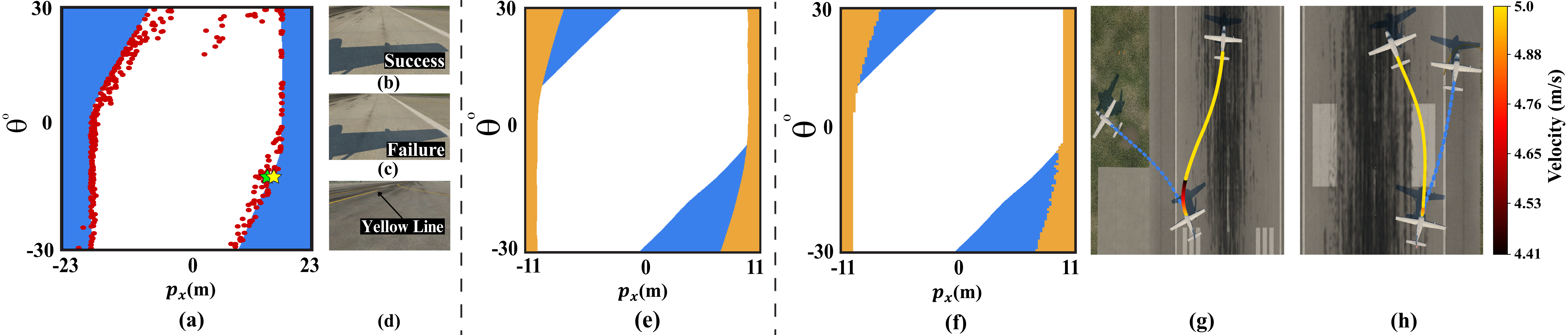}
\caption{\small{\textbf{(a)} Red dots represent incorrect predictions of the FD. The incorrect predictions are concentrated around the \NRT\ boundary. 
\textbf{(b)} Input image corresponding to the green star in (a), incorrectly classified as a failure by FD. \textbf{(c)} Input image corresponding to the yellow star in (a), incorrectly classified as a success by FD. \textbf{(d)} Test runway contains yellow-colored lines that confuse the TaxiNet, but such failures are not predicted well, as no runway with yellow lines is present in the training dataset.\textbf{(e)} \NRT\ under GPS-based state estimation fallback mechanism (yellow) and \NRT\ under TaxiNet controller (blue). \textbf{(f)} System \NRT\ under default TaxiNet controller (blue) and \NRT\ under the safety pipeline (yellow). The \NRT\ obtained using the FD and the fallback controller is appreciably smaller than the \NRT\ under TaxiNet. \textbf{(g, h)} Trajectory followed by the aircraft under the TaxiNet controller (dashed blue line) and the safety pipeline (yellow line). The color shift in the yellow curve shows velocity variation due to the fallback controller.}}
\vspace{-1em}
\label{fig:all_fbc}
\end{figure*}

For example, in Fig. \ref{fig:all_fbc}(b) and \ref{fig:all_fbc}(c), we show two visually similar images in our test dataset (corresponding to the green and yellow stars in Fig. \ref{fig:all_fbc}(a)).

Even though the two images are visually similar, one image is inside the \NRT , which leads to system failure (hence a failure), while the other is not. 
Such similar images with minor differences are hard for our FD to detect. Finally, we noticed that some of the semantics of the test environment that causes failures are not present in the training dataset and are not predicted well by the proposed FD (Fig. \ref{fig:all_fbc}(d)). Such issues highlight the need for a continual update of FDs as more data about the system failures is obtained.

\subsection{Fallback Mechanism}

Using FD, one can detect if an observation acquired by \VBC\ results in a system failure. This signal can be used to trigger a fallback mechanism that steers the system to safety when it encounters such failure situations. Essentially, we use FD as an online switch when the system encounters failures,
\begin{equation}
u = \begin{cases}
                                \policy(\image) & \quad \text { if } FD(\image) = 0 \\
                                \policy_{fallback} & \quad \text { otherwise }
                            \end{cases}
\label{eqn:fallback}
\end{equation}

where $u$ is the control to be applied on the system when it receives an observation $\image$, $\policy$ is the system's \VBC\, while $\policy_{\text{fallback}}$ is the fallback controller that we wish to design. Such controllers are often designed to ensure the system's safety even while trading off performance. These algorithms could include system takeover by a human expert or leading the system to pre-defined safe recovery sets. 
\begin{remark}
    Designing a fallback controller is a complex task and remains an active area of research. In this work, we propose two examples of fallback mechanisms for our TaxiNet case study. However, exploring the design principles of fallback mechanisms in greater depth would be a valuable direction for future work.
\end{remark}

\noindent\textbf{Fallback Controllers for TaxiNet \VBC}

\subsubsection{Fallback using additional sensors}

Here we demonstrate a fallback mechanism that assumes access to additional sensors for maintaining system safety. Systems such as autonomous vehicles are often equipped with multiple sensors, such as LiDARs and cameras, to allow for redundant information, which provides better environmental feedback. Similarly, we assume that our system has access to some noisy variant of its true state estimate, which could be acquired through sensors, such as a Global Positioning System (GPS). Whenever the FD flags an input observation as a failure the fallback is triggered, and we obtain the state estimation from GPS, which is used for control via the subsequent controller module; otherwise, the default VBC will act on the system. 
 For TaxiNet, we assume that the aircraft has access to a noisy variant of states $p_x$ and $\theta$ whenever the FD fags an input image as a failure. Thus, the fallback controller for this case is obtained as,
\begin{align*}
    \policy_{\text{fallback}} = tan&(-0.74p_{x,GPS} - 0.44 \theta_{GPS} ), \\
    (p_{x,GPS}, \theta_{GPS}) &= (p_x, \theta) + \epsilon, \quad \epsilon \sim N(\mathbf{0}, \mathbf{I})
\end{align*}
where $(p_{x,GPS}, \theta_{GPS})$ are the noisy GPS-based state observations while $(p_x, \theta)$ are the true states.

Here, we simulate the noisy signals as the true states with white noise $\epsilon$ (sampled from independent standard normal distributions). The \NRT\ obtained under such a fallback controller is shown in Fig. \ref{fig:all_fbc}(e). It shows significant reduction of failure states effectively demonstrating the influence of a reliable state estimator (the GPS in this case) for the taxiing task.

\subsubsection{Fallback using high-level control inputs}

In addition to the heading rate ($\dot \theta$), we now assume that we can directly command the system's velocity $v$ as another control signal. 

If at any point in its trajectory, the aircraft observes an image $\image$ that is classified as a failure (i.e., $FD(I) = 1$), the linear velocity of the plane ($v$ in Eqn. \eqref{eqn:dyn_taxinet}) is reduced by $0.01 m/s$. This incremental reduction in speed is applied each time a failure is detected, effectively slowing the aircraft until it observes a safe image. Upon observing a safe image, the default TaxiNet controller resumes control, or if failures continue to be detected, the aircraft may ultimately be brought to a complete stop.

Fig. \ref{fig:all_fbc}(g,h) shows two cases demonstrating the fallback controller in action using the proposed velocity modification scheme. In the first case (Fig. \ref{fig:all_fbc}(g)), the failure is triggered due to the aircraft approaching too close to the runway boundary, while in the second case (Fig. \ref{fig:all_fbc}(h)) the failure is triggered due to semantic failure of the runway markings. In both cases, the default TaxiNet controller leads the system off the runway (blue-dashed trajectories), while the fallback controller decreases the aircraft velocity whenever a failure is triggered to ensure system safety (yellow trajectories). 

We also compute the system \NRT\ under the Taxinet controller and the proposed fallback controller pipeline (Fig.\ref{fig:all_fbc}(f)).
The \NRT\ is much smaller under the safety pipeline (the \NRT\ volume decreases from 25.75\% to 14.83\%), showing a \textbf{42\% reduction} in the number of closed-loop system failures over the default controller.

We were able to run the system with the safety mechanism (using the FD and fallback controller) at 25Hz, compared to 33Hz, with the default TaxiNet governing the system as the sole controller. This allows us to run the system online without any severe latency issues.

\section{Offline Improvement via Incremental Training}
\label{sec:fallback}
The approach involving a coupled FD and fallback controller introduces an \textit{online} method to mitigate failures, enabling the preemptive detection of issues during system runtime and the implementation of preventive measures to ensure safety. In contrast, this section presents a complementary, \textit{offline} approach to enhance system safety through incremental training by reinforcing the controller against its known failure modes. These failure modes can be effectively identified using the mining process through \NRT\ computation, allowing the creation of an augmented failure dataset.
Hence, similar to the dataset obtained for the FD, these failure inputs represent scenarios where the system fails to operate optimally. Thus, retraining the controller on these specific failure instances can strengthen the system's robustness in handling such cases. 

\begin{figure*}[ht!]
\centering
\includegraphics[width=\textwidth]{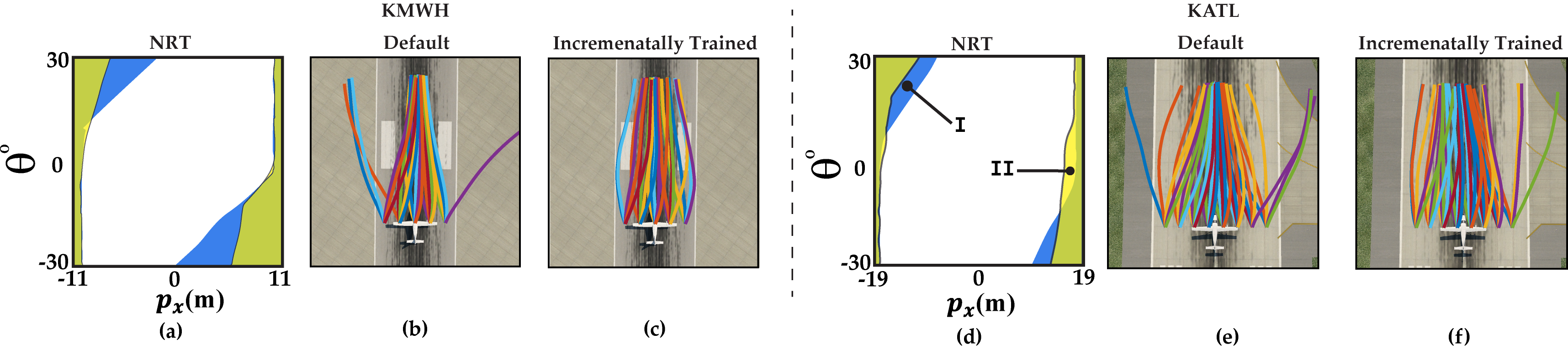}
\vspace{-1.75em}
\caption{\small{\textbf{(a)} \NRT\ slices for $p_y = 110m$, $\envp_1 = $ \texttt{morning} 
 on KMWH runway (training dataset), with the incremental controller \NRT\ (green) overlaid on the default controller \NRT\ (blue). Trajectories on the KMWH runway under \textbf{(b)} default controller and \textbf{(c)} incrementally trained controller. The incrementally trained controller shows significantly fewer failures than the default version. \textbf{(d)} \NRT\ slices for $p_y = 110m$, $\envp_1 = $ \texttt{morning}  on KATL runway, with the incremental controller \NRT\ (green) overlaid on the default controller \NRT\ (blue). (I) marks an area where the incremental controller does better than the default version, while (II) marks an area where the default controller is better. \textbf{(e)} Trajectories on the KATL runway with default controller show multiple failure trajectories. \textbf{(f)} Trajectories on the KATL runway with incrementally trained controller showing failure due to effects not seen in the augmented dataset (eg. another runway merging from the right).}}
 \vspace{-1em}
\label{fig:incremental}
\end{figure*}

\noindent \textbf{\textit{Incremental Training of the TaxiNet \VBC.}} To gather the dataset for incrementally training the Deep Neural Network (DNN) involved in the TaxiNet \VBC, we first mine the failure images from the unsafe regions of the \NRT s for morning and night in clear conditions from the KMWH runway. The labels corresponding to these images are the ground truth states, $p_x \text{ and } \theta$ from where these images are rendered in the X-Plane simulator. These data samples are added to the publicly available dataset (used for training the original TaxiNet's DNN) to create an augmented failure dataset. Keeping all the hyperparameters the same, we \textit{retrain the pre-trained TaxiNet's DNN} on the newly augmented failure dataset. Fig. \ref{fig:incremental} shows the results of incrementally training the DNN with the failure dataset. The reduction in the \NRT s for the Taxninet with the incrementally trained DNN gives evidence of the improved performance of this method. Overall, we see a \textbf{20\% decrease} in the \NRT\ volume due to incremental training. We even noted a \textbf{10\% reduction} in the DNN's prediction error over 125 random sampled trajectories in the test runway, KATL. This shows that incremental training has \textit{improved the network in both the safety and performance aspects of the taxiing task}.

Even though incremental training provided us with an improved controller, a few caveats about the algorithm need to be addressed. It's important to acknowledge that incremental training demands a deeper understanding of the base network's training process, such as hyperparameters and the dataset structure, i.e., the inputs and labels format. Another key requirement of this approach is to maintain a well-balanced failure dataset containing both safe and failure inputs. This balance ensures that the existing performance of the system is not compromised while incorporating additional training signals from the failure inputs. In our case, we had to upsample the number of failure images before augmenting them in the dataset. This ensured the number of failure samples was sufficient to guide the retraining process while not causing the network to forget what it previously knew. This symptom is often called catastrophic forgetting and is quite common in methods such as incremental training or continual learning \cite{kirkpatrick2017overcoming}. An example is seen in Fig. \ref{fig:incremental}(d), where (II) marks the section of the \NRT\ where the incrementally trained controller does worse than the default version.

Another necessary requirement of the augmented failure dataset structure is that it should be similar to the dataset used to train the original system. In our case, we were able to access the original dataset as well as the simulator used to obtain the ground truth labels, which, unfortunately, might not be available for general systems. Beyond dataset preparations, we need access to a trainable model, which might not always be publicly available. 

On the positive side, once the network is retrained, it can be utilized to identify new failure modes, allowing for further iterative retraining.  We were able to achieve impressive performance with a single iteration of incremental training. Multiple iterations of this process can potentially lead to the development of a highly robust controller. 

\section{Discussion and Future Work}
\label{sec:conclusion}
We present a framework for automatically detecting and mitigating closed-loop failures in \VBC s by integrating photorealistic simulation tools with HJ reachability analysis techniques. We utilize DeepReach to compute \NRT s across a range of state and environmental conditions, such as varying illumination throughout the day and different weather conditions. This approach leverages data-driven methods to approximate reachable tubes, thereby overcoming the computational limitations associated with grid-based methods, particularly for high-dimensional systems.

Our framework leverages the identified failure modes to achieve two main objectives: (1) implementing an online safety mechanism through an FD paired with a fallback controller and (2) employing an offline procedure to enhance the robustness of \VBC s via incremental training. We provide coverage guarantees on the recall metric of the FD to ensure a safety-first approach. The effectiveness of our method in improving system safety is demonstrated using the well-established TaxiNet controller.

Our framework has a few limitations that we would like to explore in future work. A key assumption for our algorithm is the availability of high-fidelity simulators capable of accurately replicating visual conditions across diverse environmental settings. It is possible to, however, replace simulated data with real-world measurements, yet such replacement entails using costly hardware systems in safety-critical scenarios, which may not always be practical. Secondly, computing \NRT s via Deepreach remains an ongoing area of research, and our future work will focus on extending the framework to more complex robotic systems, such as self-driving vehicles and manipulators, by incorporating multi-modal sensory inputs for enhanced control. Lastly, while the current failure detection (FD) system effectively uses a simple classifier and fallback controller, future studies will investigate integrating natural language processing to express the reasons behind a failure, leading to more effective fallback controllers and improved semantic mining of failure data.

\bibliographystyle{IEEEtran}
\bibliography{./references, ./references2}

\vfill

\end{document}